\pdfoutput=1

\documentclass[11pt]{article}

\usepackage[preprint]{acl}

\usepackage{times}
\usepackage{latexsym}
\usepackage{booktabs}
\usepackage{arydshln}
\usepackage{geometry}
\usepackage{array}
\usepackage{enumitem}
\usepackage{tikz}
\usepackage[most]{tcolorbox}

\usepackage[T1]{fontenc}

\usepackage[utf8]{inputenc}

\usepackage{microtype}

\usepackage{inconsolata}

\usepackage{graphicx}

\title{Turn-Level Empathy Prediction Using Psychological Indicators}

\author{
 \textbf{Shaz Furniturewala\textsuperscript{1}\thanks{Work done during internship at NUS Center for Trusted Internet and Community}},
 \textbf{Kokil Jaidka\textsuperscript{2}},
\\
 \textsuperscript{1}Birla Institute of Technology and Science, Pilani, \\
 \textsuperscript{2}NUS Center for Trusted Internet and Community, National University of Singapore,
\\
}

\begin{document}
\maketitle
\begin{abstract}
For the WASSA 2024 Empathy and Personality Prediction Shared Task, we propose a novel turn-level empathy detection method that decomposes empathy into six psychological indicators: Emotional Language, Perspective-Taking, Sympathy and Compassion, Extroversion, Openness, and Agreeableness. A pipeline of text enrichment using a Large Language Model (LLM) followed by DeBERTA fine-tuning demonstrates a significant improvement in the Pearson Correlation Coefficient and F1 scores for empathy detection, highlighting the effectiveness of our approach. Our system officially ranked 7th at the CONV-turn track.
\end{abstract}

\section{Introduction}
Empathy, a critical construct in human social interaction, involves perceiving, understanding, and resonating with the emotional states and perspectives of others. This construct is essential in domains such as mental health support, customer service, and human-computer interaction \cite{Paiva2014EmotionMF}. Empathy comprises both cognitive and affective dimensions \cite{batsonempathy,singerempathy}, with the cognitive dimension involving the intellectual understanding of another's psychological state and the affective dimension involving the emotional experience of another's feelings. \\
Empathy detection in natural language processing (NLP) focuses on identifying and quantifying empathetic expressions in text. Accurate empathy detection can enhance the performance of automated systems in responding to human emotions appropriately \cite{shum2018eliza}. Applications include therapeutic conversational agents, customer service bots, and social robotics. \\
Traditional empathy detection methods rely on lexical and syntactic analysis \cite{algosforempathy}, using features such as sentiment polarity, emotion lexicons, and dialogue acts. These methods often fail to capture the nuanced and context-dependent nature of empathy, rooted in deeper psychological constructs. Effective empathy detection requires a sophisticated analytical framework to interpret underlying psychological indicators. \\
Our study employs GPT-4o \cite{openai2024gpt4} to evaluate six psychological indicators for each utterance in our dataset. GPT-4o's advanced language understanding and generation capabilities allow it to assess and articulate the presence of these indicators, providing ratings and explanatory sentences. These enriched inputs are used to train a DeBERTa classifier \cite{he2021deberta}, known for its superior performance in NLP tasks due to its enhanced attention mechanisms and optimized representation learning.  Our empirical analysis demonstrates that incorporating the psychological indicators significantly enhances the performance of the empathy detection models, as evidenced by improvements in Pearson correlation, F1 scores, and accuracy metrics. \\
\setlength{\dashlinedash}{0.5pt}
\setlength{\dashlinegap}{1.5pt}

\begin{table*}[ht]
    \centering
    \begin{tabular}{c|>{\raggedright\arraybackslash}p{0.7\linewidth}|c}
    \toprule
         Speaker&  Utterance& Empathy\\ \midrule
         Person 1&  What did you think about this article& 0.6667
\\ \cdashline{1-3}
         Person 2&  It's definitely really sad to read, considering everything they're all going through. What did you think?& 4.3333
\\ \cdashline{1-3}
         Person 1&  I think it's super sad... they seem to never catch a break, always struggling.& 4.6667
\\ \cdashline{1-3}
         Person 2&  I can't imagine just living in an area that is constantly being ravaged by hurricanes or earthquakes. I take my location for granted.& 4.6667
\\ \bottomrule
    \end{tabular}
    \caption{A snippet of the dataset.}
    \label{tab:dataset}
\end{table*}
\noindent By integrating psychological indicators and leveraging advanced NLP models, our work offers a new pipeline for multi-task learning that relies on the cognitive underpinnings of human behavior. This contributes to developing more contextually aware and empathetic conversational agents, improving human-computer interaction, and enabling more emotionally intelligent automated systems.
\section{Related Work}
Simple approaches to empathy detection have employed rule-based systems and manually crafted features, leveraging predefined empathy-related keywords and patterns to identify empathetic expressions \cite{algosforempathy}. With the advent of machine learning, statistical models such as support vector machines (SVMs) and random forests were utilized, which leveraged a broader set of features, including syntactic structures, word embeddings, and discourse markers \cite{chen2020automated, mathur2021modeling}. However, these approaches are limited by their reliance on predefined patterns and surface-level features, which may not generalize well across different contexts and fail to capture the complexity and contextual nature of empathetic language. \\
Recent advancements in deep learning have further propelled the field, with neural network architectures such as recurrent neural networks (RNNs) \cite{tavabi2019multimodal}, LSTMs \cite{tan2019multimodal}, and transformers \cite{guda2021empathbert} demonstrating significant improvements in capturing the contextual dependencies and semantic richness of empathetic language. Transformer-based models, particularly BERT \cite{devlin2019bert} and its variants, have shown remarkable performance in various NLP tasks, including empathy detection. These models are expected to provide a more nuanced understanding of empathetic expressions by leveraging self-attention mechanisms that model long-range dependencies and contextual relationships within the text. Nevertheless, these models would still need sufficient context to interpret empathetic behavior and, by themselves, do not offer a way to consider the specific psychological constructs that underpin empathetic behavior. \\
To address this gap, our approach for empathy detection focused on first enriching the data with more psychological indicators and then improving upon the design of the current best-performing model. Our objective was to focus on the psychological indicators underpinning empathy, leveraging our prior work in modeling cognitive appraisals of happiness ~\cite{liu2023psyam} and purchase behavior ~\cite{yeo2023peace,yeo2024beyond} by translating a text classification paradigm into a multi-task classification problem. Our prior work successfully demonstrated the effectiveness of using psychological constructs to enhance predictive models, providing a foundation for our current approach. We believed that applying a similar framework to empathy detection would yield robust and interpretable models capable of capturing the nuanced and multifaceted nature of empathetic expressions in language.

\section{Method}
We decomposed the concept of "Empathy" into theory-inspired fundamental components of empathetic behavior \cite{batsonempathy, singerempathy}, focusing on six distinct psychological indicators:
\begin{itemize}[noitemsep]
    \item Emotional Language: Represented by the use of emotion-laden words (e.g., "sad," "happy," "worried") and descriptions of feelings or emotions, both personal and those of others.
    \item Perspective-Taking: Indicated by statements that show an understanding of another person’s point of view.
    \item Sympathy and Compassion: Demonstrated by expressions of concern for another person’s well-being.
    \item Extroversion: Reflected by signs of sociability, such as mentions of interactions with others, excitement about social events, or enjoyment of group activities.
    \item Openness: Indicated by signs of creativity, intellectual curiosity, or unconventional thinking, such as discussing diverse topics, exploring different perspectives, or expressing interest in novel ideas.
    \item Agreeableness: Shown by kindness, altruism, or cooperation in the text, for example, expressions of concern for others' well-being, willingness to help, or avoiding conflict.
\end{itemize}
By enriching our data with information about these indicators, we aimed to provide a more comprehensive and interpretable framework for empathy detection. However, the challenge remains in accurately operationalizing these psychological constructs and ensuring that models can reliably differentiate and interpret these indicators within varied contexts and expressions of empathy. To do so, we closely relied on how these concepts are defined, worded, and measured in surveys to human participants.
\subsection{Dataset}
We use the dataset created by \cite{omitaomu2022empathic} and provided as part of the WASSA 2024 Shared Task 2 \cite{giorgi2024findings,barriere2023findings}, which is an empathetic conversation dataset consisting of conversations in response to news articles. It consists of 500 conversations between AMT workers reacting to 100 articles about negative events from \cite{buechel2018modeling}. Each conversation is greater than 15 turns. This conversation data has been third-person annotated at the turn-level on a range of 0-5 for the level of empathy displayed in the text. Scores in the dataset, however, are also fractional, presumably due to averaging among reviewers. For training, we round the scores to the nearest integer; however, while computing the Pearson Correlation Coefficient, we use the original scores. We divide the 11059 utterances into a training set of 8294 and a test set of 2765. Table \ref{tab:dataset} contains a snippet of the Dataset. The results demonstrated in this paper are unofficial, based on the test dataset we created using a subset of the training data. The official result is also provided in Section 4.
\subsection{Enrichment}
For each utterance in the dataset, we used GPT-4o to detect the level of the psychological indicators described previously, rating them as Low, Medium, or High. Additionally, GPT-4o provided a sentence explaining the rating, highlighting words or phrases contributing to a high or low rating. Table \ref{tab:qualitative} in Appendix A presents the generated ratings and explanations for a sample utterance from the dataset. The prompt used to generate them is also provided in Fig 1 in Appendix A. These ratings and explanations were used as additional context and concatenated to the original utterance. Subsequently, a DeBERTa V3 classifier was trained on this new set of inputs.
\subsection{Models and baselines}
We tested two classification models: DeBERTa-v3-Large finetuned and GPT-4o for zero-shot classification. For each model, we tested its performance on just the utterance and the utterance combined with additional context provided by the six psychological indicators. For the DeBERTa classifier, we concatenated the rating and explanation for each indicator to the original utterance, separated by [SEP] tokens. For GPT-4o, we crafted an instructional prompt, providing all the information in bullet points.
\section{Results and Discussion}

\begin{table*}[!ht]
    \centering
    \resizebox{\textwidth}{!}{
    \begin{tabular}{ll|ccc}
    \toprule
        Model & Input & Pearson R & F1 (Rounded) & Accuracy (Rounded)\\ \midrule
        GPT-4o & Utterance Only & 0.38 & 0.24 & 0.29 \\
        GPT-4o & Utterance + Indicators & 0.41 & 0.20 & 0.26 \\
        DeBERTa & Utterance Only & 0.65 & 0.32 & 0.52\\
        DeBERTa & Utterance + Indicators & \textbf{0.68} & \textbf{0.35} & \textbf{0.55} \\ \bottomrule
    \end{tabular}
    }
    \caption{The Pearson Correlation Coefficient and F1 scores for each of the four classification methods.}
    \label{tab:results}
\end{table*}
\begin{table}[!ht]
\centering
\begin{tabular}{lc}
\toprule
\textbf{Psych. Indicator}        & \textbf{Pearson R} \\ \midrule
Emotional Language      & 0.481*          \\
Perspective-Taking      & 0.186*          \\
Sympathy and Compassion & 0.437*          \\
Extroversion            & -0.152*         \\
Openness                & 0.010           \\
Agreableness            & 0.120*          \\ \bottomrule
\end{tabular}
\caption{The Pearson Correlation Coefficient of the GPT-4o scores (converted to integers) for each psychological indicator with the annotated Empathy Rating.}
\label{tab:correlation}
\end{table}
In Table \ref{tab:correlation}, we demonstrate the Pearson Correlation Coefficient of the psychological indicator scores with the annotated empathy ratings. We converted the levels predicted by GPT-4o (Low, Medium, High) to integers (-1, 0, 1) and computed the coefficients of each feature with the empathy ratings. \\
Emotional Language exhibited the highest positive correlation (0.481$^{*}$), underscoring the significance of emotion-laden words in conveying empathy. Sympathy and Compassion also showed a strong positive correlation (0.437$^{*}$), validating the role of compassionate expressions in empathetic communication. Perspective-Taking had a moderate positive correlation (0.186$^{*}$), suggesting that understanding another person's point of view contributes to empathy but is less influential than direct emotional expressions. Interestingly, Extroversion had a negative correlation (-0.152$^{*}$), implying that sociability may not align with empathetic responses in these conversations. Openness showed a very weak correlation (0.010), indicating minimal impact on empathy perception, while Agreeableness had a modest positive correlation (0.120$^{*}$), reflecting a mild association with empathetic responses through expressions of kindness and cooperation. \\
Table \ref{tab:results} reports the performance of the four classification methods in terms of their Pearson Correlation Coefficient, Accuracy, and F1 Score at empathy detection. While the coefficient is computed with the fractional empathy labels, the accuracy and F1 score are computed after rounding those labels to the nearest integer. This rounding is done because 6-label classification (0 to 5 in increments of 1) is much simpler than 16-label classification (0 to 5 in increments of 0.33). Further, we found that the rounded labels have a correlation coefficient of 0.96 with the original labels, demonstrating minimal knowledge loss. There is a large gulf between the performance of GPT-4o and the trained DeBERTa classifier for both input formats, reiterating the necessity of relying on attention mechanisms for interpreting implicit concepts like empathy from dialogic data. \\
The results demonstrate the efficacy of incorporating psychological indicators into empathy detection models. The baseline DeBERTa model trained on utterances alone achieved a Pearson correlation of 0.65, an F1 score of 0.32, and an accuracy of 0.52. When augmented with the additional context from the six psychological indicators, the model's performance improved, achieving a Pearson correlation of 0.68, an F1 score of 0.35, and an accuracy of 0.55. This indicates that the enriched input provides more comprehensive information, allowing the model to understand and predict empathy levels. \\
In contrast, the zero-shot classification using GPT-4o showed more modest improvements. The model's Pearson correlation increased from 0.38 to 0.41 when augmented with the psychological indicators, although the F1 score and accuracy slightly decreased. This suggests that while the additional context benefits GPT-4o, the model may require further fine-tuning to leverage the enriched input fully.\\
It is to be noted that these results are on the test set generated as a subset of the provided training data. We were unable to conduct these experiments on the official test data as the labels for that were not available to us. The Pearson correlation coefficient for the Utterance + Indicators DeBERTa classifier on the official test dataset is 0.534.
\section{Discussion and Conclusion}
Our method for empathy detection relied on enriching the available data with more psychological indicators that could help support the ultimate Empathy label. We show that our approach boosts performance and provides interpretable AI insights, which can be crucial for applications requiring transparency and trust. \\
The significant improvement observed with the DeBERTa model underscores the importance of considering psychological components in empathy detection. However, it is the role and performance of GPT-4o, the leading LLM, that provides the most interesting insights. On the one hand, the nuanced explanations provided by GPT-4o offer valuable context that enhances the model's ability to detect empathy. On the other hand, its relatively poor performance in empathy prediction indicates that LLMs cannot yet effectively extract and understand all the underlying information in a dialogic exchange, even in a few-shot manner, highlighting the need for further exploration of reasoning-based approaches. 

\section{Limitations}
We acknowledge one particular limitation of our work. The reliance on GPT-4o for both enriching the data and attempting to label it may lead to concept drift, where the interpretation of the labels relies heavily on prompt sensitivity and adherence, and ultimately digresses from the original definition. To address this, our future work will involve obtaining additional expert annotations and conducting a thorough inspection of GPT-4o's reasoning. This will ensure that our data enrichment is valid, nuanced, and accurately reflects human intent.

\section{Acknowledgements}
This work is supported by the Ministry of Education, Singapore under its MOE AcRF TIER3 Grant (MOE-MOET32022-0001). The travel grant for this research is supported by the Department of Communication and New Media at the National University of Singapore.

\bibliography{custom}

\appendix
\section{Appendix}
Table \ref{tab:qualitative} contains an instance of the dataset and the ratings and explanations provided by GPT-4o on the six psychological indicators.\\
Figure 1 displays the prompt used to generate the ratings and explanations from GPT-4o.

\setlength{\dashlinedash}{0.5pt}
\setlength{\dashlinegap}{1.5pt}
\setlength{\arrayrulewidth}{0.5pt}

\begin{table*}[!t]
    \centering
    \begin{tabular}{>{\raggedright}m{0.2\textwidth}|>{\raggedright\arraybackslash}m{0.75\textwidth}}
    \toprule
        Utterance & You've nailed it. Bringing children into the mix definitely amplifies the desire for 
        privacy and protection. It's all about creating a safe and stable environment for them. \\ 
    \midrule
        Emotional Language & \textbf{Medium}. The text contains words like 'desire' and implies concern for children's 
        well-being, but lacks varied emotion-laden language. \\ 
    \cdashline{1-2}
        Perspective-Taking & \textbf{High}. It acknowledges the point of view of someone concerned about privacy 
        and protection for children, representing strong perspective-taking. \\ 
    \cdashline{1-2}
        Sympathy and Compassion & \textbf{High}. The concern for creating a safe and stable environment for children shows 
        significant sympathy and compassion towards their well-being. \\ 
    \cdashline{1-2}
        Extroversion & \textbf{Low}. There is no mention of social interactions, excitement about social events, 
        or enjoyment of group activities, indicating low extroversion. \\ 
    \cdashline{1-2}
        Openness & \textbf{Low}. The text does not display signs of creativity, intellectual curiosity, or 
        unconventional thinking, which reflect low openness. \\ 
    \cdashline{1-2}
        Agreeableness & \textbf{Medium}. It shows some concern for others but does not explicitly demonstrate 
        strong kindness, altruism, or cooperative behavior. \\ 
    \bottomrule
    \end{tabular}
    \caption{Ratings and explanations provided by GPT-4o for a sample utterance from the dataset.}
    \label{tab:qualitative}
\end{table*}

\begin{figure*}
\begin{tcolorbox}[colback=gray!10, 
                  width=\textwidth,
                  colframe=black, 
                  sharp corners, 
                  boxsep=5pt, 
                  left=2pt,right=2pt,top=2pt,bottom=2pt, 
                  ]
I will provide you with a text. You have to rate the text as LOW, MEDIUM, or HIGH for each of the following five psychological dimensions and explain your score in a single sentence. \\
Psychological Dimensions: \\
- Emotional Language: HIGH emotional language contains emotion-laden words (e.g., "sad," "happy," "worried") and descriptions of feelings or emotions, both their own and those of others. LOW emotional lanugage would NOT have this. \\
- Perspective-Taking: HIGH perspective taking statements indicate an understanding of another person’s point of view. LOW perspective taking statements DO NOT. \\
- Sympathy and Compassion: HIGH sympathy and compassion is represented by statements showing concern for another person’s well-being. LOW sympathy and compassion is represented by statements that DO NOT. \\
- Extroversion: HIGH extroversion is indicated by signs of sociability, such as mentioning interactions with others, excitement about social events, or enjoyment of group activities. \\
- Openness: HIGH extroversion is indicated by signs of creativity, intellectual curiosity, or unconventional thinking. This might include discussing diverse topics, exploring different perspectives, or expressing interest in novel ideas. LOW openness would not include these. \\
- Agreeableness: HIGH agreeableness is indicated by kindness, altruism, or cooperation in the text. For example, expressions of concern for others' well-being, willingness to help, or avoiding conflict. LOW agreeableness is not indicated by these. \\
Report the result in JSON format with the text, the ratings, and the single sentence explanation for each of the five psychological dimensions.
The JSON string should have the keys 'Text', 'Scores', 'Explanations'. \\
Be very strict while giving ratings and don't give HIGH ratings unnecessarily. Also, highlight factors that contributed both positively and negatively to your rating in the single sentence explanation. \\
Here is the text: [TEXT]
\end{tcolorbox}
\label{fig:prompt}
\caption{Prompt given to GPT-4o.}
\end{figure*}

\end{document}